\documentclass{article} % For LaTeX2e
\usepackage{iclr2020_conference,times}

\usepackage{amsmath,amsfonts,bm}

\def\eqref#1{equation~\ref{#1}}
\def\1{\bm{1}}

\DeclareMathAlphabet{\mathsfit}{\encodingdefault}{\sfdefault}{m}{sl}
\SetMathAlphabet{\mathsfit}{bold}{\encodingdefault}{\sfdefault}{bx}{n}

\usepackage{microtype}
\usepackage{hyperref}
\usepackage{url}

\usepackage{subfig}
\usepackage{microtype}
\usepackage{graphicx}
\usepackage{caption}
\usepackage{booktabs} % for professional tables

\usepackage{algorithm}
\usepackage{algorithmic}
\usepackage{lipsum}
\usepackage{xcolor}
\usepackage{amssymb,amsmath}

\newtheorem{definition}{Definition}
\newtheorem{lemma}{Lemma}

\newtheorem{assumption}{Assumption}

\newtheorem{proofoflemma}{Proof of Lemma}

\graphicspath{{./Images/}}
\title{Distributed Learning: Sequential Decision Making in Resource-Constrained Environments}

\author{Udari Madhushani \\
Princeton University\\
\texttt{udarim@princeton.edu} \\
\And
Naomi Ehrich Leonard \\
Princeton University\\
\texttt{naomi@princeton.edu}
}

\iclrfinalcopy % Uncomment for camera-ready version, but NOT for submission.
\begin{document}

\maketitle 

\begin{abstract}
We study cost-effective communication strategies that can be used to improve the performance of distributed learning systems in resource-constrained environments. 
For distributed learning in sequential decision making, we propose a new cost-effective partial communication protocol. We illustrate that with this protocol the group obtains the same order of performance that it obtains with full communication. Moreover, we prove that under the proposed partial communication protocol the communication cost is $O(\log T)$, where $T$ is the time horizon of the decision-making process. This improves significantly on protocols with full communication, which incur a communication cost that is $O(T)$.  We validate our theoretical results using numerical simulations.

\end{abstract}

%%%%%%%%%%%%%%%%

\section{Introduction}

In resource-constrained environments, the difficulty in constructing and maintaining large-scale infrastructure limits the possibility of developing a centralized learning system that has access to global information, resources for effectively processing that information, and the capacity to make all the decisions. Consequently, developing cost-efficient distributed learning systems, i.e., groups of units that collectively process information and make decisions with minimal resource, is an essential step towards making machine learning practical in such constrained environments. In general, most distributed learning strategies allow individuals to make decisions using locally available information \citep{kalathil2014decentralized,landgren2016distributed, madhushani2019heterogeneous}, i.e., information that they observe or is communicated to them from their neighbors. However, the performance of such systems is strongly dependent on the underlying communication structure. Such dependence inherently leads to a trade-off between communication cost and performance. Our goal is to develop high performance distributed learning systems with minimal communication cost.

We focus on developing cost-effective distributed learning techniques for sequential decision making under stochastic outcomes. Our work is motivated by the growing number of real-world applications such as clinical trials, recommender systems, and user-targeted online advertising. For example, consider a set of organizations networked to recommend educational programs to online users under high demand. Each organization makes a series of sequential decisions about which programs to recommend according to the user feedback \citep{warlop2018fighting,feraud2018decentralized}. As another example, consider a set of small pharmaceutical companies conducting experimental drug trials \citep{tossou2016algorithms,Durand2018ContextualBF}. Each company makes a series of sequential decisions about the drug administration procedure according to the observed patient feedback. In both examples, feedback received by the decision maker is stochastic, i.e., the feedback is associated with some uncertainty. This is due to the possibility that at different time steps online users (patients) can experience the same program (drug) differently due to internal and external factors, including their own state of mind and environmental conditions. 

Performance of distributed learning in these systems can be significantly improved by establishing a communication network that facilitates \textit{full communication}, whereby each organization shares all feedback immediately with others. However, communication can often be expensive and time-consuming. Under full communication, the amount of communicated data is directly proportional to the time horizon of the decision-making process. In a resource-constrained environment, when the decision-making process continues for a long time, the full communication protocol becomes impractical. We address this problem by proposing a \textit{partial communication} strategy that obtains the \emph{same order of performance} as the full communication protocol, while using a \emph{significantly smaller amount of data communication}. 

To derive and analyze our proposed strategy, we make use of the bandit framework, a mathematical model that has been developed to model sequential decision making under stochastic outcomes \citep{laiRobbins,robbins1952some}. Consider a group of agents (units) making sequential decisions in an uncertain environment. Each agent is faced with the problem of repeatedly choosing an option from the same fixed set of options \citep{kalathil2014decentralized,landgren2016distributed,landgren2016distributedCDC,landgren2020distributed,martinez2019decentralized}. After every choice, each agent receives a numerical reward drawn from a probability distribution associated with its chosen option. The objective of each agent is to maximize its individual cumulative reward, thereby contributing to maximizing the group cumulative reward.

The best strategy for an agent in this situation is to repeatedly choose the optimal option, i.e., the option that provides the maximum average reward. However, agents do not know the expected reward values of the options. Each individual is required to execute a combination of \textit{exploiting actions}, i.e., choosing the options that are known to provide high rewards, and \textit{exploring actions}, i.e., choosing the lesser known options in order to identify options that might potentially provide higher rewards. 

This process is sped up through distributed learning that relies on agents exchanging their reward values and actions with their neighbors \citep{madhushani2019heterogeneous,madhushani2020dynamic}. The protocols in these works determine when an agent observes (samples) the reward values and actions of its neighbors.  
Our proposed protocol instead determines only when an agent shares (broadcasts). A key result is that this seemingly altruistic action (sharing) provably benefits the individual as well as the group. \cite{torney2011signalling} present how this can be an evolutionarily stable strategy in animal social groups.

We define \textit{exploit-based communication} to be  information sharing by agents only when they execute exploiting actions. Similarly, we define \textit{explore-based communication} to be information sharing by agents only when agents execute exploring actions. Thus, for information sharing, we have that
\begin{align*}
\textit{full communication} =\textit{exploit-based communication} + \textit{explore-based communication}.
\end{align*}
We propose a new partial communication protocol that uses only explore-based communication. We illustrate that explore-based communication obtains the same order of performance as full communication, while incurring a significantly smaller communication cost. 

\paragraph{Key Contributions}
In this work, we study cost-efficient, information-sharing, communication protocols in sequential decision making. Our contributions include the following:
\begin{itemize}
    \item We propose a new cost-effective partial communication protocol for distributed learning in sequential decision making. The communication protocol determines information sharing.
    \item We illustrate that with this protocol the group obtains the same order of performance as it obtains with full communication. 
    \item We prove that under the proposed partial communication protocol, the communication cost is  $O(\log T)$, where $T$ is the number of decision making steps; whereas under full communication protocols, the communication cost is $O(T)$.
\end{itemize}

\paragraph{Related Work}
Previous works \citep{kalathil2014decentralized,landgren2016distributed,landgren2016distributedCDC,landgren2018social,landgren2020distributed,martinez2019decentralized} have considered the distributed bandit problem without a communication cost.  
\cite{landgren2016distributed,landgren2016distributedCDC,landgren2020distributed} use a running consensus algorithm to update estimates and provide graph-structure-dependent performance measures that predict the relative performance of agents and networks. \cite{landgren2020distributed} also address the case of a constrained reward model in which agents that choose the same option at the same time step receive no reward.  \cite{martinez2019decentralized} 
propose an accelerated consensus procedure in the case that agents know the spectral gap of the communication graph and design a decentralized UCB algorithm based on delayed rewards. 
\cite{szorenyi2013gossip} consider a P2P communication where an agent is only allowed to communicate with two other agents at each time step. In \cite{chakraborty2017coordinated}, at each time step, agents decide either to sample an option or to broadcast the last obtained reward to the entire group. In this setting, each agent suffers from an opportunity cost whenever it decides to broadcast the last obtained reward. A communication strategy where agents observe the rewards and choices of their neighbors according to a leader-follower setting is considered in \cite{landgren2018social}.  Decentralized bandit problems with communication costs are considered in the works of  \cite{tao2019collaborative, Wang2020Distributed}. \cite{tao2019collaborative} consider the pure exploration bandit problem with a communication cost equivalent to the number of times agents communicate. \cite{Wang2020Distributed} propose an algorithm that achieves near-optimal performance with a communication cost equivalent to the amount of data transmitted.  \cite{madhushani2020dynamic} propose a communication rule where agents observe their neighbors when they execute an exploring action.

%%%%%%%%%%%%%%%%

\section{Methodology}
\subsection{Problem Formulation}
In this section we present the mathematical formulation of the problem. Consider a group of $K$ agents faced with the same $N$-armed bandit problem for $T$ time steps. In this paper we use the terms arms and options interchangeably. Let $X_i$ be a sub-Gaussian random variable with variance proxy $\sigma_i^2$, which denotes the reward of option $i\in \{1,2,\ldots,N\}.$ Define $\mathbb{E}\left(X_i\right)=\mu_i$ as the expected reward of option $i.$ We define the option with maximum expected reward as the optimal option $i^*=\arg\max \{\mu_1,\ldots, \mu_N\}.$ Let $\Delta_i=\mu_{i^*}-\mu_i$ be the expected reward gap between option $i^*$ and option $i.$ Let $\mathbb{I}_{\{\varphi_t^k=i\}}$ be the indicator random variable that takes value 1 if agent $k$ chooses  option $i$ at time $t$ and 0 otherwise. 

We define the communication network as follows. Let $G(\mathcal{V},\mathcal{E})$ be a fixed nontrivial graph that defines neighbors, where $\mathcal{V}$ denotes the set of agents and $e(k,j)\in \mathcal{E}$ denotes the communication link between agents $k$ and $j.$ Let $\mathbb{I}^t_{\{\cdot,k\}}$ be the indicator variable that takes value 1 if agent $k$ shares its reward value and choice with its neighbors at time $t.$ Since agents can send reward values and choices only to their neighbors, it holds that $\mathbb{I}^t_{\{j,k\}}=0, \forall k,j,t,$ such that $e(j,k)\notin \mathcal{E}.$ 

\subsection{Our Algorithm}
Let $\widehat{\mu}^k_i(t)$ be the estimated mean of option $i$ by agent $k$ at time $t.$ Let $n_i^k(t)$ and $N^k_i(t)$ denote the number of  samples of option $i$ and the number of observations of option $i$, respectively, obtained by agent $k$ until time $t$.  $N_i^k(t)$ is equal to $n_i^k(t)$ plus the number of observations of option $i$ that agent $k$ received from its neighbors until time $t.$ So, by definition
\begin{align*}
n_i^k(t)=\sum_{\tau=1}^t\mathbb{I}_{\{\varphi_{\tau}^k=i\}},\:\:\:N_i^k(t)=\sum_{\tau=1}^t\sum_{j=1}^K\mathbb{I}_{\{\varphi_{\tau}^j=i\}}\mathbb{I}^{\tau}_{\{k,j\}}.
\end{align*}
\begin{assumption} \label{asm:Iitialization}
\normalfont
Initially, all the agents are given a reward value for one sample from each option.
\end{assumption}
The initial given reward values are used as the empirical estimates of the mean values of the options. Let $X_i^k(0)$ denote the reward  received initially by agent $k$ for option $i$. 
% For each option, each agent constructs an objective function that captures the trade-off between exploring and exploiting. At each time step agents choose the option that maximizes the objective function. 
The estimated mean value is calculated by taking the average of the total reward observed for option $i$ by agent $k$ until time $t$: 
\begin{align*}
\widehat{\mu}^k_i(t)=\frac{S_i^k(t)+X^k_i(0)}{N^k_i(t)+1}
\end{align*}
where $S_i^k(t)=\sum_{\tau=1}^t\sum_{j=1}^k X_i\mathbb{I}_{\{\varphi^j_{\tau}=i\}}\mathbb{I}^{\tau}_{\{k,j\}}.$

The goal of each agent is to maximize its individual cumulative reward, thereby contributing to maximizing the group cumulative reward. We assume known variance proxy as follows.
\begin{assumption}\label{asm:KnownVariance}
\normalfont
All agents know the variance proxy $\sigma_i^2$ of the rewards associated with each option.
\end{assumption}

\begin{assumption}
\normalfont
When more than one agent chooses the same option at the same time they receive rewards independently drawn from the probability distribution associated with the chosen option.
\end{assumption}

To realize the goal of maximizing cumulative reward, agents are required to minimize the number of times they sample sub-optimal options. Thus, each agent employs an agent-based strategy that captures the trade-off between exploring and exploiting by constructing an objective function that strikes a balance between the estimation of the expected reward and the uncertainty associated with the estimate \citep{auer2002finite}. Each agent samples options according to the following rule.

\begin{definition} {\bf{(Sampling Rule)}}\label{def:samplerule}
The sampling rule $\{\varphi^k_t\}_1^{T}$ for agent $k$ at time $t \in \{1, \ldots, T\}$ is 
    \begin{align*}
    \mathbb{I}_{\{\varphi^k_{t+1}=i\}}=\left\{
    \begin{array}{cl} 1 &, \:\:\:i=\arg \max \{Q^k_1(t),\cdots,Q^k_N(t)\}\\
      0 &, \:\:\: {\mathrm{o.w.}}\end{array}\right.
    \end{align*}
    with     
    \begin{align*}
    Q^k_i(t)\triangleq \widehat{\mu}^k_{i}(t)+C^k_i(t),\:\:\:\:\:
    C^k_i(t)\triangleq\sigma_{i}\sqrt{\frac{2(\xi+1)\log \left(t\right)}{N^k_{i}(t)+1}},\:\:\:\: \text{and} \:\:\:\:\xi>1.
    \end{align*}
    % where $\xi>1.$
\end{definition}
$C_i^k(t)$ represents agent $k$'s uncertainty of the estimated mean of option $i$. When the number of observations of option $i$ is high, the uncertainty associated with the estimated mean of option $i$ will be low; this is reflected in the inverse relation between $C_i^k(t)$ and $N_i^k(t)$.

An exploiting action corresponds to choosing the option with maximum estimated mean value.  This occurs when the option with maximum objective function value is the same as the option with maximum estimated mean value. An exploring action correspond to choosing an option with high uncertainty.  This occurs when the option with maximum objective function value is different from the option with maximum estimated mean value. Each agent can reduce the number of samples it takes from sub-optimal options by leveraging communication to reduce the uncertainty associated with the estimates of sub-optimal options. Thus, in resource-constrained environments, it is desirable to communicate reward values obtained from sub-optimal options only. Exploring actions often lead to taking samples from sub-optimal options. So, we define a partial communication protocol such that agents share their reward values with their neighbors only when they execute an exploring action. 

\begin{definition} {\bf{(Communication Rule)}}\label{def:comrule}
The communication rule for agent $k$ at time $t \in \{1, \ldots, T\}$ is 
    \begin{align*}
    \mathbb{I}^{t+1}_{\{\cdot,k\}}=\left\{
    \begin{array}{cl} 1 &, \:\:\:\varphi_{t+1}^k\neq\arg \max \{\widehat{\mu}^k_1(t),\cdots,\widehat{\mu}^k_N(t)\}\\
      0 &, \:\:\: {\mathrm{o.w.}}\end{array}\right.
    \end{align*}
\end{definition}

%%%%%%%%%%%%%%%%
\section{Results}

The goal of maximizing  cumulative reward is equivalent to minimizing cumulative regret, which is the loss incurred by the agent when sampling sub-optimal options. We analyze the performance of the proposed algorithm using expected cumulative regret and expected communication cost. 

For a group of $K$ agents facing the $N$-armed bandit problem for $T$ time steps, the expected group cumulative regret can be expressed as
\begin{align*}
\mathbb{E}\left(R(T)\right)=\sum_{i=1}^N\sum_{k=1}^K\Delta_i\mathbb{E}\left(n^k_i(T)\right).
\end{align*}
Thus, minimizing the expected group cumulative regret can be achieved by minimizing the expected number of samples taken from sub-optimal options.

\paragraph{Communication Cost}
Since communication refers to agents sharing their reward values and actions with their neighbors,  each communicated message has the same length. We define communication cost as the total number of times the agents share their reward values and actions during the decision-making process. Let $L(T)$ be the group communication cost up to time $T.$ Then, we have that
\begin{align}
L(T)=\sum_{k=1}^K\sum_{t=1}^T\mathbb{I}^{t}_{\{\cdot, k\}}. \label{eq:comCost}
\end{align}
Under full communication, expected communication cost is $O(T).$ We now proceed to analyze the expected communication cost under the proposed partial communication protocol. 

\begin{lemma}\label{lem:comcost}
Let $\mathbb{E}(L(T))$ be the expected cumulative communication cost of the group under the communication rule given in Definition \ref{def:comrule}. Then, we have that
\begin{align*}
\mathbb{E}(L(T))=O(\log T).
\end{align*}
\end{lemma}

The proof of Lemma \ref{lem:comcost} follows from Lemma 3 in the paper \cite{madhushani2020dynamic}. A detailed proof is provided in Appendix \ref{sec:App}.

\paragraph{Experimental Results}
We provide numerical simulation results illustrating the performance of the proposed sampling rule and the communication rule. For all the simulations presented in this section, we consider a group of 100 agents ($K=100$) and 10 options ($N=10$) with Gaussian reward distributions. We let the expected reward value of the optimal option be 11, the expected reward of all other options be 10, and the variance of all options be 1. We let the communication network graph $G$ be complete. We provide results with 1000 time steps ($T=1000$) using 1000 Monte Carlo simulations with $\xi=1.01$. 

\begin{figure}[h]%
    \centering
    \subfloat[Expected cumulative group regret of 100 agents with sampling rule from Definition \ref{def:samplerule} under full communication, 
    explore-based communication, exploit-based communication
    and no communication. ]{{\includegraphics[width=0.47\textwidth]{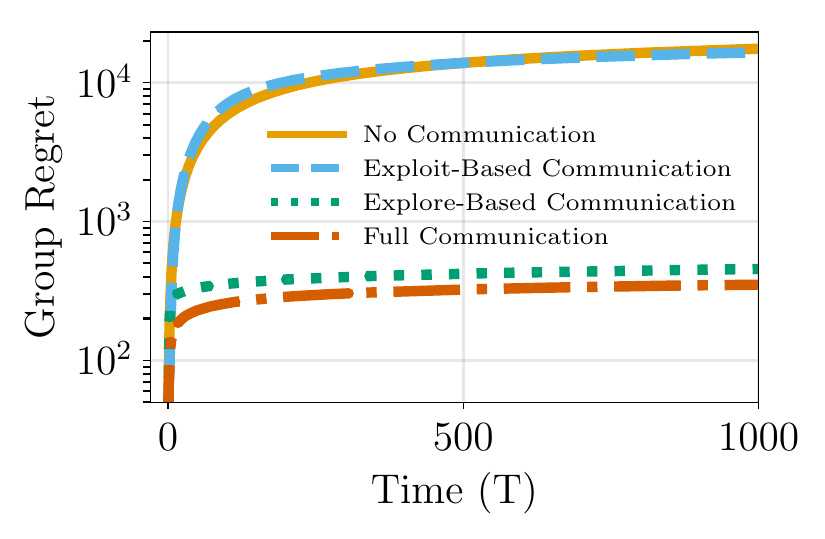}}}%
    \qquad
    \subfloat[Expected cumulative communication cost per agent for a group of 100 agents under full communication,
    explore-based communication, exploit-based communication
    and no communication.]{{\includegraphics[width=0.47\textwidth]{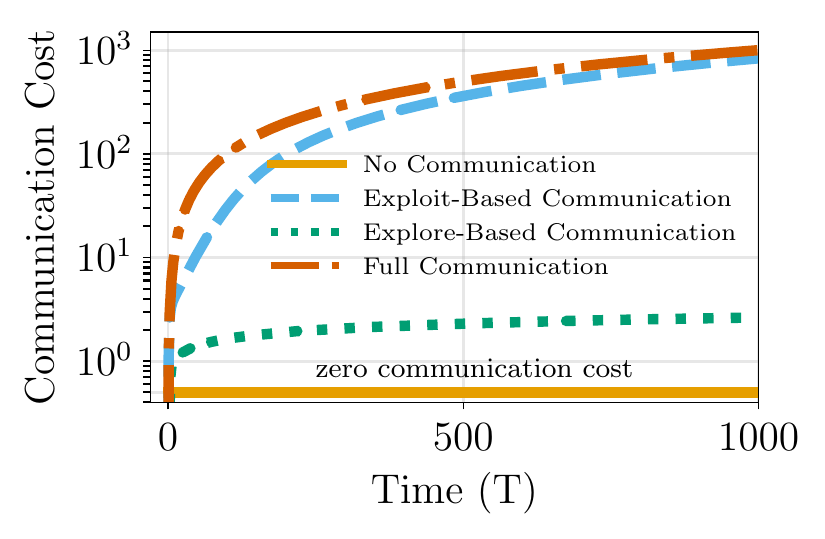}}}%
    \caption{Performance of a group of 100 agents using the sampling rule given in Definition \ref{def:samplerule} under different communication protocols.}%
    \label{fig:regretandcost}%
\end{figure}

Figure \ref{fig:regretandcost}(a) presents expected cumulative group regret for 1000 time steps. The curves illustrate that both full communication and explore-based communication significantly improve the performance of the group as compared to the case of no communication.  Further, group performance with explore-based communication is of the same order as group performance with full communication. Group performance improvement obtained with exploit-based communication is insignificant as compared to the case of no communication. 
Figure \ref{fig:regretandcost}(b) presents the results for expected cumulative communication cost per agent for 1000 time steps. The curves illustrate that communication cost incurred by explore-based communication is significantly smaller than the cost incurred by full communication and by exploit-based communication. In fact, the  cost incurred by exploit-based communication is quite close to the cost incurred by full communication. 
Overall, the results illustrate that our proposed explore-based communication protocol, in which agents share their reward values and actions only when they execute an exploring action, incurs only a small communication cost while significantly improving group performance. 

%%%%%%%%%%%%%%%%

\section{Discussion and Conclusion}

The development of cost-effective communication protocols for distributed learning is desirable in resource-constrained environments. We proposed a new partial communication protocol for sharing (broadcasting) in the distributed multi-armed bandit problem. We showed that the proposed communication protocol has a significantly smaller communication cost as compared to the case of full communication while obtaining the same order of performance. An important future extension of our work is to analyze and improve the performance of the proposed communication protocol under random communication failures.

\section*{Acknowledgement}
This research has been supported in part by ONR grants N00014-18-1-2873 and N00014-19-1-2556 and ARO grant W911NF-18-1-0325.

\bibliography{iclr2020_conference}
\bibliographystyle{iclr2020_conference}

\appendix

\section{Expected Communication Cost}\label{sec:App}

\setcounter{lemma}{0}
\begin{lemma}
Let $\mathbb{E}(L(T))$ be the expected cumulative communication cost of the group under the communication rule given in Definition \ref{def:comrule}. Then, we have that
\begin{align*}
\mathbb{E}(L(T))=O(\log T).
\end{align*}
\end{lemma}

\begin{proofoflemma}
\normalfont

The expected communication cost can be given as
\begin{align*}
\mathbb{E}\left(L(T)\right)= \sum_{k=1}^{K}\sum_{t=1}^T\mathbb{P}\left(\varphi_k^t=i,\widehat{\mu}_i^k(t)\neq \max\{\widehat{\mu}_{1}^{k}(t),\cdots,\widehat{\mu}_{N}^{k}(t)\}\right).
\end{align*}

To analyze the expected number of exploring actions, we use
\begin{align*}
    \mathbb{P}\left(\varphi_k^t=i,\widehat{\mu}_i^{k}(t)\neq \max\{\widehat{\mu}_{1}^{k}(t),\cdots,\widehat{\mu}_{N}^{k}(t)\}\right) &=  \mathbb{P}\left(\varphi_k^t=i^*,\widehat{\mu}_{i^*}^{k}(t)\neq \max\{\widehat{\mu}_{1}^{k}(t),\cdots,\widehat{\mu}_{N}^{k}(t)\}\right)\\
      + \mathbb{P}&\left(\varphi_k^t= i,\widehat{\mu}_i^{k}(t)\neq \max\{\widehat{\mu}_{1}^{k}(t),\cdots,\widehat{\mu}_{N}^{k}(t)\},i\neq i^*\right).
\end{align*}
We first upper bound the expected number of times agent $k$ broadcasts rewards and actions to its neighbors until time $T$ when it samples a sub-optimal option: 
\begin{align}
\sum_{t=1}^T\mathbb{P}\left(\varphi_k^t=i,\widehat{\mu}_{i}^{k}(t)\neq \max\{\widehat{\mu}_{1}^{k}(t),\cdots,\widehat{\mu}_{N}^{k}(t)\},i\neq i^*\right)    \leq \sum_{t=1}^T\sum_{i=1}^N\mathbb{E}\left(\mathbb{I}_{\{\varphi_k^t= i\}}\right) = O(\log T). \label{eq:subOpt}
\end{align}
This follows from the fact that we only sample sub-optimal options logarithmically with time (see Lemma 2 in \cite{madhushani2020dynamic}).

Next we analyze the expected number of times agent $k$ broadcasts rewards and actions to its neighbors until time $T$ when it samples the optimal option. Note that $\forall i,k,t$ we have
\begin{align*}
\{\varphi_k^t=i^*,\widehat{\mu}_{i^*}\neq \max\{\widehat{\mu}_{i}^{k}(t),\cdots,\widehat{\mu}_{N}^{k}(t)\}\}\subseteq  \{\widehat{\mu}_{i^*}^{k}(t)\leq \mu_{i^*}-C_{i^*}^{k}(t)\}\\
\cup\{\widehat{\mu}_{i^*}^{k}(t)\geq  \mu_{i^*}-C_{i^*}^{k}(t),\exists i \; s.t. \; \widehat{\mu}_{i}^k(t)\geq \widehat{\mu}_{i^*}^k(t),\varphi_k^t=i^*\}.\\
\end{align*}
Thus, we have
\begin{align}
\sum_{i=1}^T\mathbb{P}\left(\varphi_k^t=i^*,\widehat{\mu}_{i^*}\neq \max\{\widehat{\mu}_{i}^{k}(t),\cdots,\widehat{\mu}_{N}^{k}(t)\}\right)\leq \sum_{i=1}^T\mathbb{P}\left(\widehat{\mu}_{i^*}^{k}(t)\leq \mu_{i^*}-C_{i^*}^{k}(t)\right)\nonumber\\
+\sum_{i=1}^T\! \mathbb{P}\left(\widehat{\mu}_{i^*}^{k}(t)\geq \mu_{i^*}-C_{i^*}^{k}(t),\exists i \; s.t. \; (\widehat{\mu}_{i}^k(t)\geq \widehat{\mu}_{i^*}^{k}(t),\varphi_k^t=i^*\right).\label{eq:subex}
\end{align}
From Lemma 1 in \cite{madhushani2020dynamic} we get
\begin{align}
\sum_{i=1}^T&\mathbb{P}\left(\widehat{\mu}_{i^*}^{k}(t)\leq \mu_{i^*}-C_{i^*}^{k}(t)\right)=O(\log T),\label{eq:tailbound}
\end{align}

Now we proceed to upper bound the second summation term of (\ref{eq:subex}). Note that for some $\beta_i^k(t)>0$ we have
\begin{align*}
\sum_{i=1}^T\mathbb{P}&\left(\widehat{\mu}_{i^*}^{k}(t)\geq \mu_{i^*}-C_{i^*}^{k}(t),\widehat{\mu}_{i}^k(t)\geq \widehat{\mu}_{i^*}^k(t),\varphi_k^t=i^*\right)\\
&\leq\sum_{i=1}^T\mathbb{P}\left(\widehat{\mu}_{i^*}^{k}(t)\geq \mu_{i^*}-C_{i^*}^{k}(t),\widehat{\mu}_{i}^k(t)\geq \widehat{\mu}_{i^*}^k(t),N_{i^*}^k(t)\leq \beta_i^k(t),\varphi_k^t=i^*\right)\\
&+\sum_{i=1}^T\mathbb{P}\left(\widehat{\mu}_{i^*}^{k}(t)\geq \mu_{i^*}-C_{i^*}^{k}(t),\widehat{\mu}_{i}^k(t)\geq \widehat{\mu}_{i^*}^k(t),N_{i^*}^k(t)> \beta_i^k(t)\right)\\
&\leq \beta_i^k(T)+\sum_{i=1}^T\mathbb{P}\left(\widehat{\mu}_{i^*}^{k}(t)\geq \mu_{i^*}-C_{i^*}^{k}(t),\widehat{\mu}_{i}^k(t)\geq \widehat{\mu}_{i^*}^k(t),N_{i^*}^k(t)> \beta_i^k(t)\right).
\end{align*}

Let $i$ be the sub-optimal option with highest estimated expected reward for agents $k$ at time $t.$ Then we have $i=\arg\max \{\widehat{\mu}_{1}^{k}(t),\cdots,\widehat{\mu}_{N}^{k}(t)\}$ and $i\neq i^*.$ If agent $k$ chooses option $i^*$ at time $t+1$ we have $Q_{i^*}^k(t)> Q_{i}^k(t).$ Thus we  have $\widehat{\mu}_{i}^{k}(t)>\widehat{\mu}_{i^*}^{k}(t)$ and $C_{i}^k(t)<C_{i^*}^k(t).$ Then
for $\beta_{i}^k(t)=\frac{8\sigma_{i^*}(\xi+1)}{\Delta^2_{i}}\log t$
we have
\begin{align*}
\sum_{i=1}^T\mathbb{P}&\left(\widehat{\mu}_{i^*}^{k}(t)\geq \mu_{i^*}-C_{i^*}^{k}(t),\widehat{\mu}_{i}^k(t)\geq \widehat{\mu}_{i^*}^k(t),N_{i^*}^k(t)> \beta_i^k(t)\right)\\
&\leq \sum_{i=1}^T\mathbb{P}\left(\widehat{\mu}_{i}^{k}(t)\geq \mu_{i^*}-C_{i^*}^{k}(t),\mu_{i^*}>\mu_i+2C_{i^*}^k(t)\right)\\
&\leq 
\sum_{t=1}^T\mathbb{P}\left(\widehat{\mu}_i^k(t)\geq \mu_i+C_i^k(t)\right)=O(\log T).
\end{align*}

The last equality follows from Lemma 1 by \cite{madhushani2020dynamic}.

Then we have
\begin{align}
&\sum_{i=1}^T\mathbb{P}\left(\widehat{\mu}_{i^*}^{k}(t)\geq \mu_{i^*}-C_{i^*}^{k}(t),\exists i \; s.t. \; (\widehat{\mu}_{i}^k(t)\geq \widehat{\mu}_{i^*}^{k}(t),\varphi_k^t=i^*\right)=O(\log T)\label{eq:optSample}.
\end{align}
The proof of Lemma \ref{lem:comcost} follows from Equations (\ref{eq:subOpt})-(\ref{eq:optSample}). 

\end{proofoflemma}

\end{document}